\documentclass[runningheads]{llncs}
\usepackage{graphicx}
\usepackage{comment}
\usepackage{amsmath,amssymb} 
\usepackage{color}

\usepackage{breqn}
\usepackage[noend, boxruled, linesnumbered] {algorithm2e}
\usepackage{enumitem}
\usepackage{booktabs}
\usepackage{hhline}

\makeatletter
\def\@seccntformat#1{\@ifundefined{#1@cntformat}%
   {\csname the#1\endcsname\quad}  
   {\csname #1@cntformat\endcsname}
}
\let\oldappendix\appendix 
\renewcommand\appendix{%
    \oldappendix
    \newcommand{\section@cntformat}{\appendixname~\thesection\quad}
}
\makeatother

\DeclareMathOperator*{\argmax}{arg\,max}
\DeclareMathOperator*{\argmin}{arg\,min}

\begin{document}
\pagestyle{headings}
\mainmatter
\def\ECCVSubNumber{6083}  

\title{Weakly Supervised Instance Segmentation by Learning Annotation Consistent Instances} 

\titlerunning{WSIS by AnnoCoIn}
%
\author{Aditya Arun\inst{1} \and
C.V. Jawahar\inst{1} \and
M. Pawan Kumar\inst{2}}
\authorrunning{A. Arun et al.}
%
\institute{CVIT, KCIS, IIIT Hyderabad \and 
OVAL, University of Oxford}
\maketitle

\begin{abstract}
Recent approaches for weakly supervised instance segmentations depend on two components: (i) a pseudo label generation model which provides instances that are consistent with a given annotation; and (ii) an instance segmentation model, which is trained in a supervised manner using the pseudo labels as ground-truth. Unlike previous approaches, we explicitly model the uncertainty in the pseudo label generation process using a conditional distribution. The samples drawn from our conditional distribution provide accurate pseudo labels due to the use of semantic class aware unary terms, boundary aware pairwise smoothness terms, and annotation aware higher order terms. Furthermore, we represent the instance segmentation model as an annotation agnostic prediction distribution. In contrast to previous methods, our representation allows us to define a joint probabilistic learning objective that minimizes the dissimilarity between the two distributions. Our approach achieves state of the art results on the PASCAL VOC 2012 data set, outperforming the best baseline by $4.2\%\ \text{mAP}^r_{0.5}$ and $4.8\%\ \text{mAP}^r_{0.75}$.
\end{abstract}

\section{Introduction}\label{sec:intro}

The instance segmentation task is to jointly estimate the class labels and segmentation masks of the individual objects in an image. Significant progress on instance segmentation has been made based on the convolutional neural networks (CNN)~\cite{chen2018masklab,he2017mask,li2017fully,liu2018path,novotny2018semi}. However, the traditional approach of learning CNN-based models requires a large number of training images with instance-level pixel-wise annotations. Due to the high cost of collecting these supervised labels, researchers have looked at training these instance segmentation models using weak annotations, ranging from bounding boxes~\cite{hsu2019weakly,khoreva2017simple} to image-level labels~\cite{ahn2019weakly,cholakkal2019object,ge2019label,laradji2019masks,zhou2018weakly,zhu2019learning}.

Many of the recent approaches for weakly supervised instance segmentation can be thought of as consisting of two components. First, a pseudo label generation model, which provides instance segmentations that are consistent with the weak annotations. Second, an instance segmentation model which is trained by treating the pseudo labels as ground-truth, and provides the desired output at test time.

Seen from the above viewpoint, the design of a weakly supervised instance segmentation approach boils down to three questions. First, how do we represent the instance segmentation model? Second, how do we represent the pseudo label generation model? And third, how do we learn the parameters of the two models using weakly supervised data? The answer to the first question is relatively clear: we should use a model that performs well when trained in a supervised manner, for example, Mask R-CNN~\cite{he2017mask}. However, we argue that the existing approaches fail to provide a satisfactory answer to the latter two questions.

Specifically, the current approaches do not take into account the inherent uncertainty in the pseudo label generation process~\cite{ahn2019weakly,laradji2019masks}. Consider, for instance, a training image that has been annotated to indicate the presence of a person. There can be several instance segmentations that are consistent with this annotation, and thus, one should not rely on a single pseudo label to train the instance segmentation model. Furthermore, none of the existing approaches provide a coherent learning objective for the two models. Often they suggest a simple two-step learning approach, that is, generate one set of pseudo labels followed by a one time training of the instance segmentation model~\cite{ahn2019weakly}. While some works consider an iterative training procedure~\cite{laradji2019masks}, the lack of a learning objective makes it difficult to analyse and adapt them in varying settings.

In this work, we address the deficiencies of prior work by (i) proposing suitable representations for the two aforementioned components; and (ii) estimating their parameters using a principled learning objective. In more detail, we explicitly model the uncertainty in pseudo labels via a conditional distribution. The conditional distribution consists of three terms: (i) a semantic class aware unary term to predict the score of each segmentation proposal; (ii) a boundary aware pairwise term that encourages the segmentation proposal to completely cover the object; and (iii) an annotation consistent higher order term that enforces a global constraint on all segmentation proposals (for example, in the case of image-level labels, there exists at least one corresponding segmentation proposal for each class, or in the case of bounding boxes, there exists a segmentation proposal with sufficient overlap to each bounding box). All three terms combined enable the samples drawn from the conditional distribution to provide accurate annotation consistent instance segmentations. Furthermore, we represent the instance segmentation model as an annotation agnostic prediction distribution. This choice of representation allows us to define a joint probabilistic learning objective that minimizes the dissimilarity between the two distributions. The dissimilarity is measured using a task-specific loss function, thereby encouraging the models to produce high quality instance segmentations.

We test the efficacy of our approach on the Pascal VOC 2012 data set. We achieve $50.9\%\ \text{mAP}^r_{0.5}$, $28.5\%\ \text{mAP}^r_{0.75}$ for image-level annotations and $32.1\%\ \text{mAP}^r_{0.75}$ for bounding box annotations, resulting in an improvement of over 4\% and 10\% respectively over the state-of-the-art.


\section{Related Work}\label{sec:related_work}

Due to the taxing task of acquiring the expensive per-pixel annotations, many weakly supervised methods have emerged that can leverage cheaper labels. For the task of semantic segmentation various types of weak annotations, such as image-level~\cite{ahn2018learning,huang2018weakly,oh2017exploiting,pinheiro2015image}, point~\cite{bearman2016s}, scribbles~\cite{lin2016scribblesup,vernaza2017learning}, and bounding boxes~\cite{dai2015boxsup,papandreou1502weakly}, have been utilized. However, for the instance segmentation, only image-level~\cite{ahn2019weakly,cholakkal2019object,ge2019label,laradji2019masks,zhou2018weakly,zhu2019learning} and bounding box~\cite{hsu2019weakly,khoreva2017simple} supervision have been explored. Our setup considers both the image-level and the bounding box annotations as weak supervision. For the bounding box annotations, Hsu \emph{et al.}~\cite{hsu2019weakly} employs a bounding box tightness constraint and train their method by employing a multiple instance learning (MIL) based objective but they do not model the annotation consistency constraint for computational efficiency. 

Most of the initial works~\cite{zhou2018weakly,zhu2019learning} on weakly supervised instance segmentation using image-level supervision were based on the class activation maps (CAM)~\cite{oquab2015object,selvaraju2017grad,wei2017object,zhou2016learning}. In their work, Zhou \emph{et al.}~\cite{zhou2018weakly} identify the heatmap as well as its peaks to represent the location of different objects. Although these methods are good at finding the spatial location of each object instance, they focus only on the most discriminative regions of the object and therefore, do not cover the entire object. Ge \emph{et al}~\cite{ge2019label} uses the CAM output as the initial segmentation seed and refines it in a multi-task setting, which they train progressively. We use the output of~\cite{zhou2018weakly} as the initial segmentation seed of our conditional distribution but the boundary aware pairwise term in our conditional distribution encourages pseudo labels to cover the entire object.

Most recent works on weakly supervised learning adopt a two-step process - generate pseudo labels and train a supervised model treating these pseudo labels as ground truth. Such an approach provides state-of-the-art results for various weakly supervised tasks like object detection~\cite{arun2019dissimilarity,tang2017multiple,tang2018weakly}, semantic segmentation~\cite{dai2015boxsup,khoreva2017simple}, and instance segmentation~\cite{ahn2019weakly,laradji2019masks}. Ahn \emph{et al.}~\cite{ahn2019weakly} synthesizes pseudo labels by learning the displacement fields and pairwise pixel affinities. These pseudo labels are then used to train a fully supervised Mask R-CNN~\cite{he2017mask}, which is used at the test time. Laradji \emph{et al.}~\cite{laradji2019masks} iteratively samples the pseudo segmentation label from MCG segmentation proposal set~\cite{APBMM2014} and train a supervised Mask R-CNN~\cite{he2017mask}. This is similar in spirit to our approach of using the two distributions. However, they neither have a unified learning objective for the two distribution nor do they model the uncertainty in their pseudo label generation model. Regardless, the improvement in the results reported by these two methods advocates the importance of modeling two separate distributions. In our method, we explicitly model the two distributions and define a unified learning objective that minimizes the dissimilarity between them.

Our framework has been inspired by the work of Kumar \emph{et al.}~\cite{kumar2012modeling} who were the first to show the necessity of modeling uncertainty by employing two separate distributions in a latent variable model. This framework has been adopted for weakly supervised training of CNNs for learning human poses and object detection tasks~\cite{arun2018learning,arun2019dissimilarity}. While their framework provides an elegant formulation for weakly supervised learning, its various components need to be carefully constructed for each task. Our work can be viewed as designing conditional and prediction distributions, as well as the corresponding inference algorithms,  which are suited to instance segmentation.

\section{Method}\label{sec:method}

\subsection{Notation}\label{ssec:notation}
We denote an input image as $\mathbf{x} \in \mathbb{R}^{(H\times W \times 3)}$, where $H$ and $W$ are the height and the width of the image respectively. For each image, a set of segmentation proposals $\mathcal{R} = \{r_1, \dots, r_P\}$ are extracted from a class-agnostic object proposal algorithm. In this work, we use Multiscale Combinatorial Grouping (MCG)~\cite{APBMM2014} to obtain the object proposals. For the sake of simplicity, we only consider image-level annotations in our description. However, our framework can be easily extended to other annotations such as bounding boxes. Indeed, we will use bounding box annotations in our experiments. Given an image and the segmentation proposals, our goal is to classify each of the segmentation proposals to one of the $C+1$ categories from the set $\{0, 1, \dots, C\}$. Here category $0$ is the background and categories $\{1, \dots, C\}$ are object classes. 

We denote the image-level annotations by $\mathbf{a} = \{0, 1\}^C$, where $\mathbf{a}^{(j)} = 1$ if image $x$ contains the $j-$th object. Furthermore, we denote the unknown instance-level (segmentation proposal) label as $\mathbf{y} = \{0, \dots, C\}^P$, where $\mathbf{y}^{(i)} = j$ if the $i-$th segmentation proposal is of the $j-$th category. A weakly supervised data set $\mathcal{W} = \{(\mathbf{x}_n, \mathbf{a}_n) \mid n = 1,\dots,N\}$ contains $N$ pairs of images $\mathbf{x}_n$ and their corresponding image-level annotations $\mathbf{a}_n$.

\subsection{Conditional Distribution}\label{ssec:cond_dist}
Given the weakly supervised data set $\mathcal{W}$, we wish to generate pseudo instance-level labels $\mathbf{y}$ such that they are annotation consistent. Specifically, given the segmentation proposals $\mathcal{R}$ for an image $\mathbf{x}$, there must exists at least one segmentation proposal for each image-level annotation $\mathbf{a}^{(j)} = 1$. Since the annotations are image-level, there is inherent uncertainty in the figure-ground separation of the objects. We model this uncertainty by defining a distribution $\Pr_c(\mathbf{y} \mid \mathbf{x}, \mathbf{a}; \boldsymbol{\theta}_c)$ over the pseudo labels conditioned on the image-level weak annotations. Here, $\boldsymbol{\theta}_c$ are the parameters of the distribution. We call this a \emph{conditional distribution}. 

The conditional distribution itself is not explicitly represented. Instead, we use a neural network with parameters $\boldsymbol{\theta}_c$ which generates samples that can be used as pseudo labels. For the generated samples to be accurate, we wish that they have the following three properties: (i) they should have high fidelity with the scores assigned by the neural network for each region proposal belonging to each class; (ii) they should cover as large a portion of an object instance as possible; and (iii) they should be consistent with the annotation. 

\subsubsection{Modeling:} In order for the conditional distribution to be annotation consistent, the instance-level labels $\mathbf{y}$ need to be compatible with the image-level annotation $\mathbf{a}$. This constraint cannot be trivially decomposed over each segmentation proposal. As a result, it would be prohibitively expensive to model the conditional distribution directly as one would be required to compute its partition function. Taking inspiration from Arun \emph{et al.}~\cite{arun2019dissimilarity}, we instead draw representative samples from the conditional distribution using the Discrete \textsc{Disco} Nets~\cite{bouchacourt2017thesis}. We will now describe how we model the conditional distribution through a Discrete \textsc{Disco} Nets, which we will now call a \emph{conditional network}.

\begin{figure}[t]
    \centering
    \includegraphics[width=0.99\textwidth]{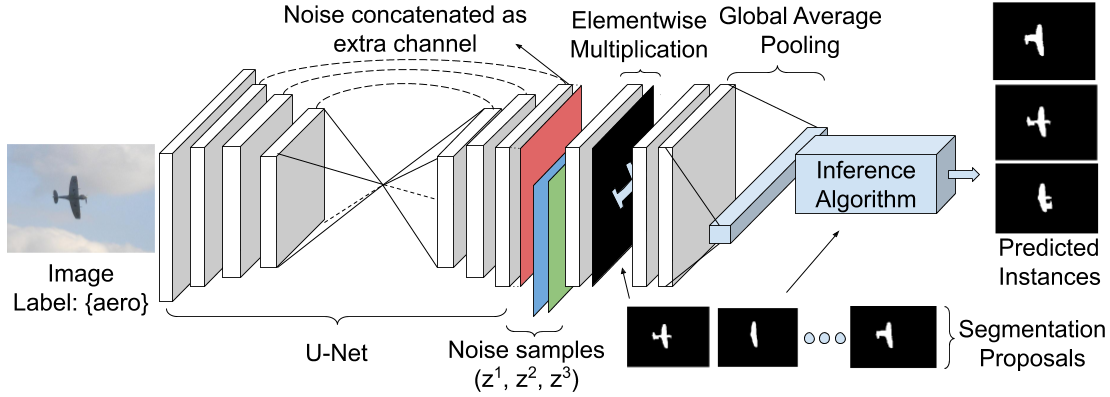}
    \caption{\emph{The conditional network: a modified U-Net architecture is used to model the conditional network. For a single input image and three different noise samples $\{z^1, z^2, z^3\}$ (represented as red, green, and blue matrix) and a pool of segmentation proposals, three different instances are predicted for the given weak annotation (aeroplane in this example). Here the noise sample is concatenated as an extra channel to the final layer of the U-Net. The segmentation proposals are multiplied element-wise with the global feature to obtain the proposal specific feature. A global average pooling is applied to get class specific score. Finally, an inference algorithm generates the predicted samples.}}
    \label{fig:cond_net}
\end{figure}

Consider the modified fully convolutional U-Net~\cite{ronneberger2015u} architecture shown in figure~\ref{fig:cond_net} for the conditional distribution. The parameters of the conditional distribution $\boldsymbol{\theta}_c$ are modeled by the weights of the conditional network. Similar to~\cite{kohl2018probabilistic}, noise sampled from a uniform distribution is added after the U-Net block (depicted by the colored filter). Each forward pass through the network takes the image $\mathbf{x}$ and noise sample $\mathbf{z}^k$ as input and produces a score function $F_{u,\mathbf{y}_u}^k(\boldsymbol{\theta}_c)$ for each segmentation proposal $u$ and the corresponding putative label $\mathbf{y}_u$. We generate $K$ different score functions using $K$ different noise samples. These  score  functions  are  then  used  to  sample  the  segmentation  region proposals $\mathbf{y}_c^k$ such that they are annotation consistent. This enables us to efficiently generate the samples from the underlying distribution.

\subsubsection{Inference:} Given the input pair $(\mathbf{x}, \mathbf{z}^k)$ the conditional network outputs $K$ score functions for each of the segmentation proposal $F_{u,\mathbf{y}_u}^k(\boldsymbol{\theta}_c)$. We redefine these score functions to obtain a final score function such that it is then used to sample the segmentation region proposals $\mathbf{y}_c^k$. The final score function has the following three properties. 
\begin{enumerate}
    \item The score of the sampled segmentation region proposal should be consistent with the score function. This \emph{semantic class aware unary term} ensures that the final score captures the class specific features of each segmentation proposal. Formally, $G_{u,\mathbf{y}_u}^k(\mathbf{y}_c) = F_{u,\mathbf{y}_u}^k(\boldsymbol{\theta}_c)$.
    \item The  unary  term  alone  is  biased  towards  segmentation proposals that are highly discriminative. This results in selecting a segmentation proposal which does not cover the object in its entirety. We argue that all the neighboring segmentation proposals must have the same score discounted by the edge weights between them. We call this condition \emph{boundary aware pairwise term}. 
    
    In order to make the score function $G_{u,\mathbf{y}_u}^k(\mathbf{y}_c)$ pairwise term aware, we employ a simple but efficient iterative algorithm. The algorithm proceeds by iteratively updating the scores $G_{u,\mathbf{y}_u}^k(\mathbf{y}_c)$ by adding the contribution of their neighbors discounted by the edge weights between them until convergence. In practice, we fix the number of iteration to $3$. Note that, it is possible to backpropagate through the iterative algorithm by simply unrolling its iterations, similar to a recurrent neural networks (RNN). Formally,
    \begin{dmath}
        G_{u,\mathbf{y}_u}^{k,n}(\mathbf{y}_c) = G_{u,\mathbf{y}_u}^{k,n-1}(\mathbf{y}_c) + \frac{1}{H_{u, v}^{k,n-1}(\mathbf{y}_c) + \delta} \exp{(-I_{u,v})}.
    \label{icm_scoring}
    \end{dmath}
    Here, $n$ denotes the iteration step for the iterative algorithm and $\delta$ is a small positive constant added for numerical stability. In our experiments, we set $\delta = 0.1$. The term $H_{u, v}^{k, n-1}(\mathbf{y}_c)$ is the difference between the scores of the neighboring segmentation proposal. It helps encourage same label for the neighboring segmentation proposals that are not separated by the edge pixels. It is given as,
    \begin{dmath}
        H_{u, v}^{k, n-1}(\mathbf{y}_c) = \sum_{u, v \in \mathcal{N}_u} \left( G_{u,\mathbf{y}_u}^{k,n-1}(\mathbf{y}_c) - G_{v,\mathbf{y}_u}^{k,n-1}(\mathbf{y}_c) \right)^2.
        \label{pairwise_computation}
    \end{dmath}
    The term $I_{u,v}$ is the sum of the edge pixel values between the two neighboring segmentation regions. Note that the pairwise term is a decay function weighted by the edge pixel values. This ensures a high contribution to the pairwise term is only from the pair of segmentation proposals that does not share an edge.
    \item In order to ensure that at there must exist at least one segmentation proposal for every image-level annotation,a higher order penalty is added to the score. We call this \emph{annotation consistent higher order term}. Formally,
    \begin{dmath}
        S^k(\mathbf{y}_c) = \sum_{u=1}^P G_{u,\mathbf{y}_u}^{k,n}(\mathbf{y}_c) + Q^k(\mathbf{y}_c).
    \label{higher_order_scoring}
    \end{dmath}
    Here,
    \begin{dmath}
        Q^k(\mathbf{y}_c) = \begin{cases}
                    0 
                        & \text{if}\ \forall j \in \{1, \dots, C\}\ \mathrm{s.t.}\ \mathbf{a}^{(j)}=1, \\
                        &\; \exists i \in \mathcal{R}\ \mathrm{s.t.}\ {\bf y}^{(i)}=j,	\\
                    -\infty & \text{otherwise}.
                \end{cases}
    \label{higher_order_constraint_scoring}
    \end{dmath}
    Given the scoring function in equation (\ref{higher_order_scoring}), we compute the $k-$th sample of the conditional network as,
    \begin{dmath}
        \mathbf{y}_c^{k} = \argmax_{\mathbf{y} \in \mathcal{Y}} S^k(\mathbf{y}_c).
    \label{objective}
    \end{dmath}
    Observe that in equation (\ref{objective}), the $\argmax$ is computed over the entire output space $\mathcal{Y}$. A na\"ive brute force algorithm is therefore not feasible. We design an efficient greedy algorithm that selects the highest scoring non-overlapping proposal. The inference algorithm is described in Algorithm~\ref{algo1:inference}.
\end{enumerate}

\begin{algorithm}[t]
\label{algo1:inference}
\RestyleAlgo{boxed}
\DontPrintSemicolon
\SetAlgoLined
\setlength{\textfloatsep}{0pt}
\caption{Inference Algorithm for the Conditional Net}
\SetKwInOut{Input}{Input}\SetKwInOut{Output}{Output}
\Input{Region masks: $R$, Image-level labels: $a$}
\Output{Predicted instance level instances: $\mathbf{y}_c^k$}
\BlankLine
\tcc{Iterative Algorithm}
$G_{u,\mathbf{y}_u}^k(\mathbf{y}_c) = F_{u,\mathbf{y}_u}^k(\boldsymbol{\theta}_c)$\;
\Repeat{$G_{u,\mathbf{y}_u}^{k,n}(\mathbf{y}_c)$ has coverged}
{
    \For{$v \in \mathcal{N}_u$}
    {
        $H_{u, v}^{k, n-1}(\mathbf{y}_c) = \sum_{u, v \in \mathcal{N}_u} \left( G_{u,\mathbf{y}_u}^{k,n-1}(\mathbf{y}_c) - G_{v,\mathbf{y}_v}^{k,n-1}(\mathbf{y}_c) \right)^2.$\;
        $G_{u,\mathbf{y}_u}^{k,n}(\mathbf{y}_c) = G_{u,\mathbf{y}_u}^{k,n-1}(\mathbf{y}_c) + \frac{1}{H_{u, v}^{k,n-1}(\mathbf{y}_c) + \delta} \exp{(-I_{u,v})}$
    }
}
\tcc{Greedily select highest scoring non-overlapping proposal}
$Y \gets \phi $ \;
\For{$j \gets \{1, \dots, C\} \wedge \mathbf{a}^{(j)} = 1$}
{
    $Y_j \gets \phi$\;
    $R_j \gets sort(G_{u,\mathbf{y}_u}^{k,n}(\mathbf{y}_c))$\;
    \For{$i \in 1, \dots, P$}
    {
        $Y_j \gets r_i$\;
        $R_j \gets R_j - r_i$\;
        \For{$l \in R_j \wedge \frac{r_i \cap r_l}{r_l} > t$}
        {
            $R_j \gets R_j - r_l$
        }
    }
    $Y \gets Y_j$
}
\Return $\mathbf{y}_c^k = Y$\;
\end{algorithm}

\subsection{Prediction Distribution}\label{sec:pred_dist}
The task of the supervised instance segmentation model is to predict the instancemask given an image. We employ Mask R-CNN [18] for this task. As predictions for each of the regions in the Mask R-CNN is computed independently, we can view the output of the Mask R-CNN as the following fully factorized distribution,
\begin{dmath}
    \Pr_p(\mathbf{y} \mid \mathbf{x};\boldsymbol{\theta}_p) = \prod_{i=1}^{R} \Pr(\mathbf{y}_i \hiderel{\mid} \mathbf{r}_i, \mathbf{x}_i; \boldsymbol{\theta}_p).
    \label{pred_dist}
\end{dmath}

Here, $R$ are the set of bounding box regions proposed by the region proposal network and $\mathbf{r}_i$ are its corresponding region features. The term $\mathbf{y}_i$ is the corresponding prediction for each of the bounding box proposals. We call the above distribution a \emph{prediction distribution} and the Mask R-CNN a \emph{prediction network}.

\section{Learning Objective}\label{sec:learning_obj}

Given the weakly supervised data set $\mathcal{W}$, our goal is to learn the parameters of the prediction and the conditional distribution, $\boldsymbol{\theta}_p$ and $\boldsymbol{\theta}_c$ respectively. We observe that the task of both the prediction and the conditional distribution is to predict the instance segmentation mask. Moreover, the conditional distribution utilizes the extra information in the form of image-level annotations. Therefore, it is expected to produce better instance segmentation masks. Leveraging the task similarity between the two distribution, we would like to bring the two distribution close to each other. Inspired by the work of~\cite{arun2018learning,arun2019dissimilarity,bouchacourt2016disco,kumar2012modeling}, we design a joint learning objective that can minimize the dissimilarity coefficient~\cite{rao1982diversity} between the prediction and the conditional distribution. In what follows, we briefly describe the dissimilarity coefficient before applying it to our setting.

\paragraph{Dissimilarity Coefficient:} The dissimilarity coefficient between any two distributions $\Pr_1(\cdot)$ and $\Pr_2(\cdot)$ is determined by measuring their diversities. Given a task-specific loss function $\Delta(\cdot, \cdot)$, the diversity coefficient between the two distribution $\Pr_1(\cdot)$ and $\Pr_2(\cdot)$ is defined as the expected loss between two samples drawn randomly from the two distributions respectively. Formally,
\begin{dmath}
    DIV_{\Delta}(\Pr\nolimits_1, \Pr\nolimits_2) = \mathbb{E}_{y_1 \sim \Pr_1(\cdot)} \left[ \mathbb{E}_{y_2 \sim \Pr_2(\cdot)}[\Delta(\mathbf{y}_1, \mathbf{y}_2)] \right].
\label{diversity_coeff}
\end{dmath}
If the model brings the two distributions close to each other, we could expect the diversity $DIV_\Delta(\Pr_1, \Pr_2)$ to be small. Using this definition, the dissimilarity coefficient is defined as the following Jensen difference,
\begin{dmath}
    DISC_\Delta(\Pr\nolimits_1, \Pr\nolimits_2) = DIV_\Delta(\Pr\nolimits_1, \Pr\nolimits_2) - \gamma DIV_\Delta(\Pr\nolimits_2, \Pr\nolimits_2) \\ 
      - (1-\gamma) DIV_\Delta(\Pr\nolimits_1, \Pr\nolimits_1),
\label{disco}
\end{dmath}
where, $\gamma = [0,1]$. In our experiments, we use $\gamma = 0.5$, which results in dissimilarity coefficient being symmetric for the two distributions.

\subsection{Task-Specific Loss Function:}\label{ssection:task_specific_loss}
The dissimilarity coefficient objective requires a task-specific loss function. To this end, we use the multi-task loss defined by Mask R-CNN~\cite{he2017mask} as,
\begin{dmath}
    \Delta(\mathbf{y}_1, \mathbf{y}_2) = \Delta_{\text{cls}}(\mathbf{y}_1, \mathbf{y}_2) + \Delta_{\text{box}}(\mathbf{y}_1, \mathbf{y}_2) + \Delta_{\text{mask}}(\mathbf{y}_1, \mathbf{y}_2).
\label{task_loss}
\end{dmath}
Here, $\Delta_{\text{cls}}$ is the classification loss defined by the log loss, $\Delta_{\text{box}}$ is the bounding box regression loss defined as the smooth-L1 loss, and $\Delta_{\text{mask}}$ is the segmentation loss for the mask defined by pixel-wise cross entropy, as proposed by~\cite{he2017mask}.

Note that the conditional network outputs the segmentation region $\mathbf{y}$, where there are no bounding box coordinates predicted. Therefore, for the conditional network, only $\Delta_{\text{cls}}$ and $\Delta_{\text{mask}}$ is active as the gradients for $\Delta_{\text{box}}$ is $0$. For the prediction network, all three components of the loss functions are active. We construct a tight bounding box around the pseudo segmentation label, which acts as a pseudo bounding box label for Mask R-CNN.

\subsection{Learning Objective for Instance Segmentation:} \label{ssection:learning_objective}
We now specify the learning objective for instance segmentation using the dissimilarity coefficient and the task-specific loss function defined above as,
\begin{dmath}
    \theta_p^\ast, \theta_c^\ast = \argmin_{\theta_p,\theta_c} DISC_\Delta \left(\Pr\nolimits_p (\theta_p), \Pr\nolimits_c (\theta_c)\right).
\label{learning_objective}
\end{dmath}
As discussed in Section \ref{ssec:cond_dist}, modeling the conditional distribution directly is difficult. Therefore, the corresponding diversity terms are computed by stochastic estimators from $K$ samples $y_c^k$ of the conditional network. Thus, each diversity term is written as\footnote{Details in Appendix A},
\begin{dgroup}
\begin{dmath}
    DIV_\Delta(\Pr\nolimits_p, \Pr\nolimits_c) = \frac{1}{K}  \sum_{k=1}^K \sum_{{\mathbf{y}}_p^{(i)}} \Pr\nolimits_p({\mathbf{y}}_p^{(i)}; \boldsymbol{\theta}_p)  \Delta({\mathbf{y}}_p^{(i)}, {\mathbf{y}}_c^{k}),
\label{cross_div_expanded}
\end{dmath}
\begin{dmath}
    DIV_\Delta(\Pr\nolimits_c, \Pr\nolimits_c) = \frac{1}{K(K-1)} \sum_{\substack{k,k'=1 \\ k' \neq k}}^K  \Delta(\mathbf{y}_c^k, {\mathbf{y}}_c^{k'}),
\label{self_div_expanded}
\end{dmath}
\begin{dmath}
    DIV_{\Delta}(\Pr\nolimits_p, \Pr\nolimits_p) =  \sum_{{\mathbf{y}}_p^{(i)}} \sum_{{\mathbf{y}'}_p^{(i)}} \Pr\nolimits_p({\mathbf{y}}_p^{(i)}; \boldsymbol{\theta}_p) \Pr\nolimits_p({\mathbf{y}'}_p^{(i)}; \boldsymbol{\theta}_p) \Delta({\mathbf{y}}_p^{(i)}, {\mathbf{y}'}_p^{(i)})
\label{self_div_prediction_expanded}
\end{dmath}
\end{dgroup}
Here, $DIV_\Delta(\Pr_p, \Pr_c)$ measures the cross diversity between the prediction and the conditional distribution, which is the expected loss between the samples of the two distribution. Since $\Pr_p$ is a fully factorized distribution, the expectation of its output can be trivially computed. As $\Pr_c$ is not explicitly modeled, we draw $K$ different samples to compute its required expectation.

\section{Optimization}\label{sec:optimization}

As the parameters of the two distribution, $\theta_p$ and $\theta_c$ are modeled by a neural network, it is ideally suited to be minimized by stochastic gradient descent. We employ a block coordinate descent strategy to optimize the two sets of parameters. The algorithm proceeds by iteratively fixing the prediction network and training the conditional network, followed by learning the prediction network for a fixed conditional network. 

The iterative learning strategy results in a fully supervised training of each network by using the output of the other network as the pseudo label. This allows us to readily use the algorithms developed in Mask R-CNN~\cite{he2017mask} and Discrete \textsc{Disco} Nets~\cite{bouchacourt2017thesis}. Note that, as the conditional network obtains samples over the $\argmax$ operator in equation (\ref{objective}), the objective (\ref{learning_objective}) for the conditional network is non-differentiable. However, the scoring function $S^k(\mathbf{y}_c)$ in equation (\ref{higher_order_scoring}) itself is differentiable. This allows us to use the direct loss minimization strategy~\cite{hazan2010direct,song2016training} developed for computing estimated gradients over the $\argmax$ operator~\cite{bouchacourt2017thesis,lorberbom2019direct}. We provide the details of the algorithm in appendix B.

\subsection{Visualization of the learning process}
\begin{figure}[t!]
    \centering
    \includegraphics[width=0.99\textwidth]{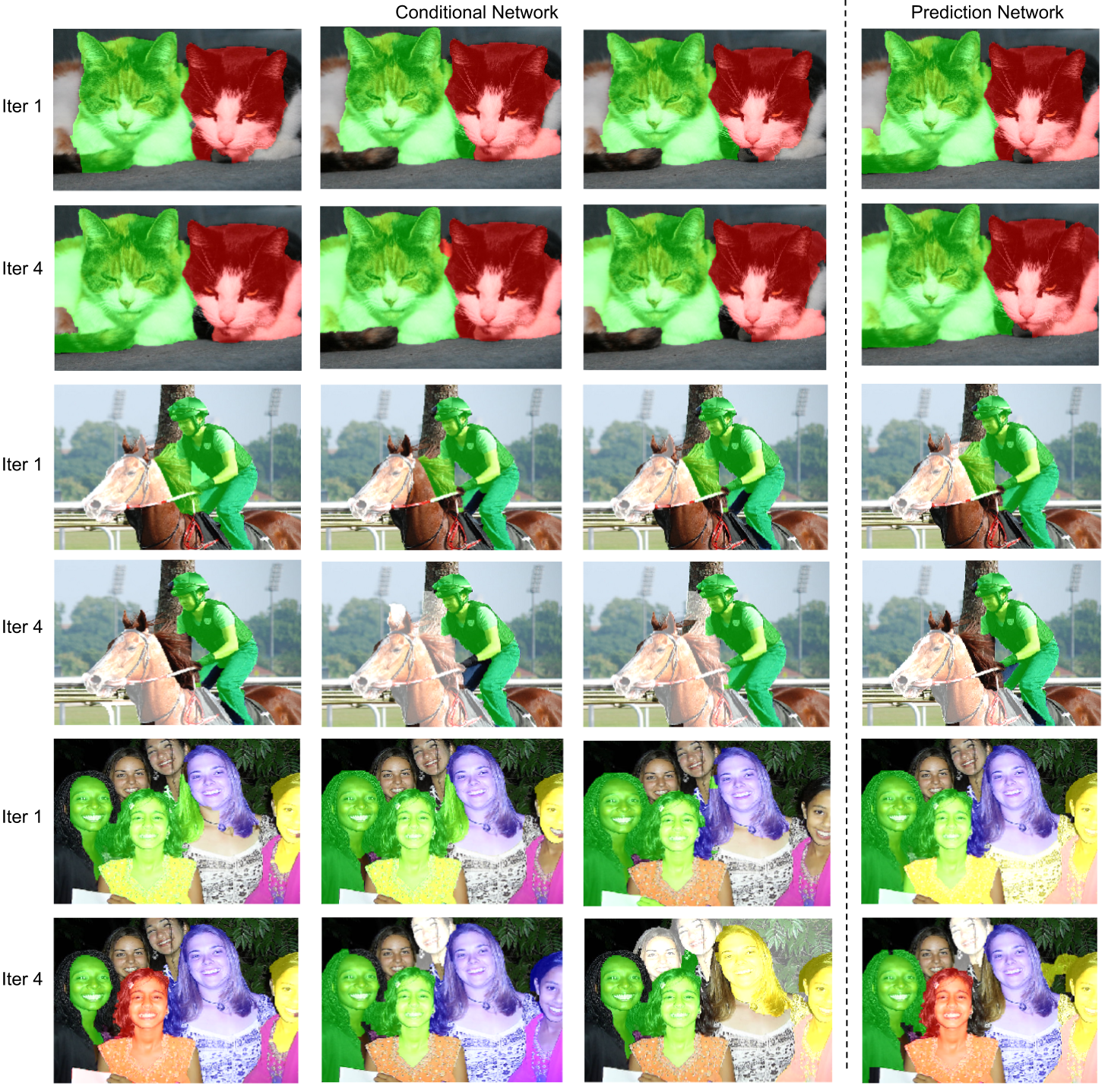}
    \caption{\emph{Examples of the predictions from the conditional and prediction networks for three different cases of varying difficulty. Columns 1 through 3 are different samples from the conditional network. For each case, its first row shows the output of the two networks after the first iteration and its second row represents the output of the two networks after the fourth (final) iteration. Each instance of an object is represented by different mask color. Best viewed in color.}}
    \label{fig:viz_learning}
\end{figure}

Figure~\ref{fig:viz_learning} provides the visualization of the output of the two networks for the first and the final iterations of the training process. The first three columns are the three output samples of the conditional distribution. Note that in our experiments, we output 10 samples corresponding to 10 different noise samples. The fourth column shows the output of the prediction distribution. The output for the prediction network is selected by employing a non-maximal suppression (NMS) with its score threshold kept at $0.7$, as is the default setting in~\cite{he2017mask}. The first row represents the output of the two networks after the first iteration and the second row shows their output after the fourth (final) iteration.

The first case demonstrates an easy example where two cats are present in the image. Initially, the conditional distribution samples the segmentation proposals which do not cover the entire body of the cat but still manages to capture the boundaries reasonably well. However, due to the variations in these samples, the prediction distribution learns to better predict the extent of the cat pixels. This, in turn, encourages the conditional network to generate a better set of samples. Indeed, by the fourth iteration, we see an improvement in the quality of samples by the conditional network and they now cover the entire body of the cat, thereby improving the performance. As a result, we can see that finally the prediction network successfully learns to segment the two cats in the image.

The second case presents a challenging scenario where a person is riding a horse. In this case, the person is occluding the front and the rear parts of the horse. Initially, we see that the conditional network only provides samples for the most discriminative region of the horse - its face. The samples generated for the person class, though not accurate, covers the entire person. We observe that over the subsequent iterations, we get an accurate output for the person class. The output for the horse class also expands to cover its front part completely. However, since its front and the rear parts are completely separated, the final segmentation could not cover the rear part of the horse.

The third case presents another challenging scenario where there are multiple people present. Four people standing in front and two are standing at the back. Here, we observe that initially, the conditional network fails to distinguish between the two people standing in the front-left of the image and fails to detect persons standing at the back. The samples for the third and the fourth persons standing in front-center and front-right respectively are also not accurate. Over the iterations, the conditional network improves its predictions for the four people standing in front and also sometimes detect the people standing at the back. As a result, prediction network finally detects five of the six people in the image.

\section{Experiments}\label{sec:experimets}

\subsection{Data set and Evaluation Metric}\label{ssec:dataset_metric}
\paragraph{Data Set:} We evaluate our proposed method on Pascal VOC 2012 segmentation benchmark~\cite{everingham2010pascal}. The data set consists of $20$ foreground classes. Following previous works~\cite{ahn2019weakly,ge2019label,hsu2019weakly,khoreva2017simple}, we use the augmented Pascal VOC 2012 data set~\cite{hariharan2011semantic}, which contains $10,582$ training images.

From the augmented Pascal VOC 2012 data set, we construct two different weakly supervised data sets. The first data set is where we retain only the image-level annotations. For the second data set, we retain the bounding box information along with the image-level label. In both the data sets, the pixel-level labels are discarded.

\paragraph{Evaluation Metric:} We adopt the standard evaluation metric for instance segmentation, mean average precision (mAP)~\cite{hariharan2014simultaneous}. Following the same evaluation protocol from other competing approaches, we report mAP with four intersection over union (IoU) thresholds, denoted by $mAP_k^r$ where $k$ denotes the different values of IoU and $k = \{0.25, 0.50, 0.70, 0.75\}$.

\subsection{Initialization}\label{ssec:initialization}
We now discuss various strategies to initialize our conditional network for different levels of weakly supervised annotations.

\paragraph{Image Level Annotations:} Following the previous works on weakly supervised instance segmentation from image-level annotations~\cite{ahn2019weakly,laradji2019masks,zhu2019learning}, we use the Class Activation Maps (CAMs) to generate the segmentation seeds for each image in the training set. Specifically, like~\cite{ahn2019weakly,laradji2019masks,zhu2019learning}, we rely on the Peak Response Maps (PRM)~\cite{zhou2018weakly} to generate segmentation seeds that identify the salient parts of the objects. We utilize these seeds as pseudo segmentation labels to initially train our conditional network. We also filter the MCG~\cite{APBMM2014} segmentation proposal such that each selected proposal has at least a single pixel overlap with the PRM segmentation seeds. This helps us reduce the number of segmentation proposals needed thereby reducing the memory requirement. Once the initial training for the conditional network is over, we proceed with the iterative optimization strategy, described in section~\ref{sec:optimization}.

\paragraph{Bounding Box Annotations} For the weakly supervised data set where bounding box annotations are present, we filter the MCG~\cite{APBMM2014} segmentation proposals such that only those who have a high overlap with the ground-truth bounding boxes are retained. The PRM~\cite{zhou2018weakly} segmentation seeds are also pruned such that they are contained within each of the bounding box annotations.

\begin{table}[t]
    \centering
    \caption{Evaluation of instance segmentation results from different methods with varying level of supervision on Pascal VOC 2012 \emph{val} set. The terms $\mathcal{F}$, $\mathcal{B}$, and $\mathcal{I}$ denotes a fully supervised approach, methods that uses the bounding box labels, and methods that uses the image-level labels respectively. Our prediction network results when using a ResNet based conditional network is presented as `Ours (ResNet-*) and the results of the prediction network using a U-Net based conditional network is presented as `Ours'. }
    \begin{tabular}{l|c|c|c|c|c|c}
            \hline
            \multicolumn{1}{c|}{Method} & Supervision & Backbone & $\text{mAP}_{0.25}^r$ & $\text{mAP}_{0.50}^r$ & $\text{mAP}_{0.70}^r$ & $\text{mAP}_{0.75}^r$ \\ \hline
            Mask R-CNN~\cite{he2017mask} & $\mathcal{F}$ & R-101 & 76.7 & 67.9 & 52.5 & 44.9 \\ \hhline{=|=|=|=|=|=|=}
            PRN~\cite{zhou2018weakly} & $\mathcal{I}$ & R-50 & 44.3 & 26.8 & - & 9.0 \\
            IAM~\cite{zhu2019learning} & $\mathcal{I}$ & R-50 & 45.9 & 28.8 & - & 11.9 \\
            OCIS~\cite{cholakkal2019object} & $\mathcal{I}$ & R-50 & 48.5 & 30.2 & - & 14.4 \\
            Label-PEnet~\cite{ge2019label} & $\mathcal{I}$ & R-50 & 49.1 & 30.2 & - & 12.9 \\
            WISE~\cite{laradji2019masks} & $\mathcal{I}$ & R-50 & 49.2 & 41.7 & - & 23.7 \\ 
            IRN~\cite{ahn2019weakly} & $\mathcal{I}$ & R-50 & - & 46.7 & - & 23.5 \\ \hline
            Ours (ResNet-50) & $\mathcal{I}$ & R-50 & 59.1 & 49.7 & 29.2 & 27.1 \\ 
            Ours & $\mathcal{I}$ & R-50 & \textbf{59.7} & \textbf{50.9} & \textbf{30.2} & \textbf{28.5} \\ \hhline{=|=|=|=|=|=|=}
            SDI~\cite{khoreva2017simple} & $\mathcal{B}$ & R-101 & - & 44.8 & -  & 46.7 \\
            BBTP~\cite{hsu2019weakly} & $\mathcal{B}$ & R-101 & \textbf{75.0} & \textbf{58.9} & 30.4 & 21.6 \\ \hline
            Ours (ResNet-101) & $\mathcal{B}$ & R-101 & 73.1 & 57.7 & 33.5 & 31.2 \\ 
            Ours & $\mathcal{B}$ & R-101 & 73.8 & 58.2 & \textbf{34.3} & \textbf{32.1} \\ \hline
        \end{tabular}
    \label{tab:comparison}
\end{table}

\subsection{Comparison with other methods}\label{ssec:comparison}
We compare our proposed method with other state-of-the-art weakly supervised instance segmentation methods. The mean average precision (mAP) over different IoU thresholds are shown in table~\ref{tab:comparison}. Compared with the other methods, our proposed framework achieves state-of-the-art performance for both image-level and the bounding box labels. We also study the effect of using a different conditional network architecture based on ResNet-50 and ResNet-101. This is shown in the table as `Ours (ResNet-50)' and `Ours (ResNet 101)' respectively. Our main result employs a U-Net based architecture for the conditional network and is presented by `Ours' in the table. The implementation details and the details of the alternative architecture are presented in appendix C. The encoder-decoder architecture of the U-Net allows us to learn better features. As a result, we observe that our method which adopts U-Net architecture for the conditional network consistently outperforms the one which adopts a ResNet based architecture. In table~\ref{tab:comparison}, observe that our approach performs particularly well for the higher IoU thresholds ($\text{mAP}_{0.70}^r$ and $\text{mAP}_{0.75}^r$) for both the image-level and the bounding-box labels. This demonstrates that our model can predict the instance segments most accurately by respecting the object boundaries. The per-class quantitative and qualitative results for our method is presented in appendix C.

\begin{table}[t]
    \centering
    \caption{Evaluation of the instance segmentation results for the various ablative settings of the conditional distribution on Pascal VOC 2012 data set}
    \begin{tabular}{c|c|c|c|c|c|c|c|c}
        \hline
        \multicolumn{3}{c|}{$\text{mAP}_{0.25}^r$} & \multicolumn{3}{c|}{$\text{mAP}_{0.50}^r$} & \multicolumn{3}{c}{$\text{mAP}_{0.75}^r$} \\ \hline
        U & U+P & U+P+H & U & U+P & U+P+H & U & U+P & U+P+H \\ \hline
        57.9 & 59.1 & 59.7 & 47.6 & 49.9 & 50.9 & 23.1 & 26.9 & 28.5 \\ \hline
    \end{tabular}
    \label{tab:ablative_terms}
\end{table}

\subsection{Ablation Experiments}
\subsubsection{Effect of the unary, the pairwise and the higher order terms} We study the effect of the conditional distributions unary, pairwise and the higher order terms have on the final output in table~\ref{tab:ablative_terms}. We use the terms U, U+P, and U+P+H to denote the settings where only the unary term is present, both the unary and the pairwise terms are present and all three terms are present in the conditional distribution. We see that unary term alone performs poorly across the different IoU thresholds. We argue that this is because of the bias of the unary term for segmenting only the most discriminative regions. The pairwise term helps allay this problem and we observe a significant improvement in the results. This is specially noticeable for higher IoU thresholds that require more accurate segmentation. The higher order term helps in improving the accuracy by ensuring that correct samples are generated by the conditional distribution. 

\begin{table}[t]
    \centering
    \caption{Evaluation of the instance segmentation results for the various ablative settings of the loss function's diversity coefficient terms on Pascal VOC 2012 data set}
    \begin{tabular}{c|c|c|c|c}
        \hline
        \begin{tabular}{@{}c@{}} Method\\ $\text{mAP}_{k}^r$ \end{tabular}  & \begin{tabular}{@{}c@{}}$\Pr_p, \Pr_c$ \\ (proposed)\end{tabular} & $PW_p, \Pr_c$ & $\Pr_p, PW_c$ & $PW_p,PW_c$ \\ \hline
        $\text{mAP}_{0.25}^r$ & 59.7 & 59.5 & 57.3 & 57.2 \\ \hline
        $\text{mAP}_{0.50}^r$ & 50.9 & 50.3 & 46.9 & 46.6 \\ \hline
        $\text{mAP}_{0.75}^r$ & 28.5 & 27.7 & 23.4 & 23.0 \\ \hline
    \end{tabular}
    \label{tab:ablative_diversity}
\end{table}

\subsubsection{Effect of the probabilistic learning objective}
To understand the effect of explicitly modeling the two distributions ($\Pr_p$ and $\Pr_c$), we compare our approach with their corresponding pointwise network. In order to sample a single output from our conditional network, we remove the self-diversity coefficient term and feed a zero noise vector (denoted by $PW_c$). For a pointwise prediction network, we remove its self-diversity coefficient. The prediction network still outputs the probability of each proposal belonging to a class. However, by removing the self-diversity coefficient term, we encourage it to output a peakier distribution (denoted by $PW_p$). Table~\ref{tab:ablative_diversity} shows that both the diversity coefficient term is important for maximum accuracy. We also note that modeling uncertainty over the pseudo label generation model by including the self-diversity in the conditional network is relatively more important. The self-diversity coefficient in the conditional network enforces it to sample a diverse set of outputs which helps in dealing with the difficult cases and in avoiding overfitting during training.

\section{Conclusion}

We present a novel framework for weakly supervised instance segmentation. Our framework efficiently models the complex non-factorizable, annotation consistent and boundary aware conditional distribution that allows us to generate accurate pseudo segmentation labels. Furthermore, our framework provides a joint probabilistic learning objective for training the prediction and the conditional distributions and can be easily extendable to different weakly supervised labels such as image-level and bounding box annotations. Extensive experiments on the benchmark Pascal VOC 2012 data set has shown that our probabilistic framework successfully transfers the information present in the image-level annotations for the task of instance segmentation achieving state-of-the-art result for both image-level and bounding box annotations.

\section{Acknowledgements}

This work is partly supported by DST through the IMPRINT program. Aditya Arun is supported by Visvesvaraya Ph.D. fellowship.

%
%
\bibliographystyle{splncs04}
\bibliography{egbib}

\clearpage


\appendix

\section{Learning Objective}
In this section we provide detailed derivation of the objective presented in the section 4.2.

Given the loss function $\Delta$ (\ref{task_loss}) which is tuned for the task of instance segmentation, we compute the diversity terms (\ref{diversity_coeff}). Recall that the diversity for any two distributions is the expected loss of the samples drawn from the two distributions. For the prediction distribution $\Pr\nolimits_p$ and the conditional distribution $\Pr\nolimits_c$, we derive the diversity between them and their self diversities as follows.

\paragraph{Diversity between prediction network and conditional network:} Following (\ref{diversity_coeff}), the diversity between prediction and conditional distribution can be written as, 
\begin{dmath}
	DIV_{\Delta}(\Pr\nolimits_p, \Pr\nolimits_c) = \mathbb{E}_{\mathbf{y}_p \sim \Pr_p(\mathbf{y} | \mathbf{x}; \boldsymbol{\theta}_p)} \big[ \mathbb{E}_{\mathbf{y}_c \sim \Pr_c(\mathbf{y} | \mathbf{x}, \mathbf{a}; \boldsymbol{\theta}_c)} [ \Delta(\mathbf{y}_p, \mathbf{y}_c) ] \big].
\end{dmath}
We then write the expectation with respect to the conditional distribution (the inner distribution) as expectation over the random variables $\mathbf{z}$ with distribution $\Pr(\mathbf{z})$ using Law of the Unconscious Statistician (LOTUS). The expectation over the random variable $\mathbf{z}$ with distribution $\Pr(\mathbf{z})$ is approximated by taking $K$ samples from $\Pr(\mathbf{z})$,
\begin{dmath}
	DIV_{\Delta}(\Pr\nolimits_p, \Pr\nolimits_c) = \mathbb{E}_{\mathbf{y}_p \sim \Pr\nolimits_p(\mathbf{y} | \mathbf{x}; \boldsymbol{\theta}_p)} \Big[ \frac{1}{K} \sum_{k=1}^K \Delta(\mathbf{y}_p, \mathbf{y}_c^{k}) \Big].
\end{dmath}
We finally compute the expectation with respect to the prediction distribution as,
\begin{dmath}
	DIV_{\Delta}(\Pr\nolimits_p, \Pr\nolimits_c) = \frac{1}{K} \sum_{k=1}^K \sum_{\mathbf{y}_p^{(i)}} \Pr\nolimits_p(\mathbf{y}_p^{(i)}; \boldsymbol{\theta}_p)  \Delta(\mathbf{y}_p^{(i)}, \mathbf{y}_c^{k}).
\end{dmath}

\paragraph{Self diversity for conditional network:} As above, using (\ref{diversity_coeff}), we write the self diversity coefficient of the conditional distribution as
\begin{dmath}
	DIV_{\Delta}(\Pr\nolimits_c, \Pr\nolimits_c)  = \mathbb{E}_{\mathbf{y}_c \sim \Pr\nolimits_c(\mathbf{y} | \mathbf{x}, \mathbf{a}; \boldsymbol{\theta}_c)} \big[ \mathbb{E}_{ \mathbf{y}_c' \sim \Pr\nolimits_c(\mathbf{y} | \mathbf{x}, \mathbf{a}; \boldsymbol{\theta}_c)} [ \Delta(\mathbf{y}_c, \mathbf{y}_c') ] \big].
\end{dmath}
We now write the two expectations with respect to the conditional distribution as the expectation over the random variables $\mathbf{z}$ and $\mathbf{z}'$ respectively. In order to approximate the expectation over the random variables $\mathbf{z}$ and $\mathbf{z}'$, we use $K$ samples from the distribution $\Pr(\mathbf{z})$ as,
\begin{dmath}
	DIV_{\Delta}(\Pr\nolimits_c, \Pr\nolimits_c)  = \frac{1}{K} \sum_{k=1}^K \frac{1}{K-1} \sum_{\substack{ k'=1, \\ k' \neq k}}^K \Delta(\mathbf{y}_c^{k}, {\mathbf{y}}_c^{k'}).
\end{dmath}
On re-arranging the above equation, we get
\begin{dmath}
	DIV_{\Delta}(\Pr\nolimits_c, \Pr\nolimits_c) = \frac{1}{K(K-1)} \sum_{\substack{k,k'=1 \\ k' \neq k}}^K  \Delta( \mathbf{y}_c^{k}, \mathbf{y}_c^{k'}).
\end{dmath}

\paragraph{Self diversity for prediction network:} Similar to the above two cases, using (\ref{diversity_coeff}), we can write the self diversity of the prediction network as
\begin{dmath}
	DIV_{\Delta}(\Pr\nolimits_p, \Pr\nolimits_p) = \mathbb{E}_{ \mathbf{y}_p \sim \Pr\nolimits_p(\mathbf{y} | \mathbf{x}; \boldsymbol{\theta}_p)} \big[ \mathbb{E}_{\mathbf{y}'_p \sim \Pr\nolimits_p(\mathbf{y} | \mathbf{x}; \boldsymbol{\theta}_p)} [ \Delta(\mathbf{y}_p, {\mathbf{y}'}_p) ] \big].
\end{dmath}
Note that the prediction distribution is a fully factorized distribution, and we can compute its exact expectation. Therefore, we compute the two expectations with respect to the inner prediction distribution as,
\begin{dmath}
     DIV_{\Delta}(\Pr\nolimits_p, \Pr\nolimits_p) = \mathbb{E}_{\mathbf{y}_p \sim \Pr\nolimits_p(\mathbf{y} | \mathbf{x}; \boldsymbol{\theta}_p)} \Big[ \sum_{{\mathbf{y}'}_p^{(i)}} \Pr\nolimits_p({\mathbf{y}'}_p^{(i)}; \boldsymbol{\theta}_p) \Delta({\mathbf{y}}_p, {\mathbf{y}'}_p^{(i)}) \Big];
\end{dmath}
and the expectation with respect to the outer prediction distribution as,
\begin{dmath}
   DIV_{\Delta}(\Pr\nolimits_p, \Pr\nolimits_p) = \sum_{{\mathbf{y}}_p^{(i)}} \sum_{{\mathbf{y}'}_p^{(i)}} \Pr\nolimits_p({\mathbf{y}}_p^{(i)}; \boldsymbol{\theta}_p) \Pr\nolimits_p({\mathbf{y}'}_p^{(i)}; \boldsymbol{\theta}_p) \Delta({\mathbf{y}}_p^{(i)}, {\mathbf{y}'}_p^{(i)}).
\end{dmath}

\section{Optimization}
\label{ap:optimization}
\subsection{Optimization over Prediction Distribution}
As parameters $\boldsymbol{\theta}_c$ of the conditional distribution are constant, the learning objective of the prediction distribution is written as,
\begin{dmath}
    \boldsymbol{\theta}_p^* = \argmin_{\boldsymbol{\theta}_p} DIV_\Delta(\Pr\nolimits_p, \Pr\nolimits_c) - (1-\gamma) DIV_\Delta(\Pr\nolimits_p, \Pr\nolimits_p).
\label{pred_opt}
\end{dmath}
This results in a fully supervised training of the Mask R-CNN network~\cite{he2017mask}. Note that the only difference between training of a standard Mask R-CNN architecture and our prediction network is the use of the dissimilarity objective function (\ref{pred_opt}) above, instead of simply minimizing the multi-task loss of the Mask R-CNN.

The prediction network takes as the input an image and the $K$ predictions sampled from the conditional network. Treating these outputs of the conditional network as the pseudo ground truth label, we compute the gradient of our dissimilarity coefficient based loss function. As the objective (\ref{pred_opt}) above is differentiable with respect to parameters $\boldsymbol{\theta}_p$, we update the network by employing stochastic gradient descent.

\subsection{Optimization over Conditional Distribution}
Similar to the prediction network, the conditional network is optimized by treating the parameters of the prediction network $\theta_p$ as constant. Its learning obective is given as,
\begin{dmath}
    \boldsymbol{\theta}_c^* = \argmin_{\boldsymbol{\theta}_c} DIV_\Delta(\Pr\nolimits_p, \Pr\nolimits_c) - \gamma DIV_\Delta(\Pr\nolimits_c, \Pr\nolimits_c).
\label{cond_opt}
\end{dmath}

\paragraph{A non-differentiable training procedure:} The conditional network is modeled using a Discrete \textsc{Disco} Nets which employs a sampling step from the scoring function $S^k(\mathbf{y}_c)$. This sampling step makes the objective function non-differentiable with respect to the parameters $\boldsymbol{\theta}_c$, even though the scoring function $S^k(\mathbf{y}_c)$ itself is differentiable. However, as the prediction network is fixed, the above objective function reduces to the one used in Bouchacourt \emph{et al.} \cite{bouchacourt2017thesis} for fully supervised training. Therefore, similar to Bouchacourt \emph{et al.}~\cite{bouchacourt2017thesis} we solve this problem by estimating the gradients of our objective function with the help of temperature parameter $\epsilon$ as,
\begin{dmath}
		\nabla_{{\boldsymbol{\theta}}_c} DISC_{\Delta}^{\epsilon}(\Pr\nolimits_p({\boldsymbol{\theta}}_p), \Pr\nolimits_c({\boldsymbol{\theta}}_c)) =  \pm \lim_{\epsilon \rightarrow 0} \frac{1}{\epsilon}\big( DIV_{\Delta}^{\epsilon}(\Pr\nolimits_p, \Pr\nolimits_c) 
        - \gamma DIV_{\Delta}^{\epsilon}(\Pr\nolimits_c, \Pr\nolimits_c)\big)
\label{eq13:gradCondNet}
\end{dmath}
where,
\begin{dmath}
		DIV_{\Delta}^{\epsilon}(\Pr\nolimits_p, \Pr\nolimits_c) = \mathbb{E}_{{\bf y}_p \sim \Pr\nolimits_p({\boldsymbol{\theta}}_p)} \left[ \mathbb{E}_{{\bf z}^k \sim \Pr({\bf z})} [ \nabla_{{\boldsymbol{\theta}}_c} S^k(\hat{\bf y}_a) - \nabla_{{\boldsymbol{\theta}}_c} S^k(\hat{\bf y}_c) ] \right],
\end{dmath}
\begin{dmath}
        DIV_{\Delta}^{\epsilon}(\Pr\nolimits_c, \Pr\nolimits_c) =  \mathbb{E}_{\mathbf{z}^k \sim \Pr({\bf z})} \left[ \mathbb{E}_{\mathbf{z}^{k'} \sim \Pr(\mathbf{z})} [ \nabla_{{\boldsymbol{\theta}}_c} S^k(\hat{\bf y}_b) - \nabla_{{\boldsymbol{\theta}}_c} S^{k'}({\hat{\bf y}'}_c) ] \right],
\label{eq14:expGradCondNet}
\end{dmath}
and,
\begin{equation}
\label{eq15:sampleGradCondNet}
	\begin{split}
		\hat{\bf y}_c & = \argmax_{{\bf y} \in \mathcal{Y}} S^k({\bf y}_c) \\
        {\hat{\bf y}'}_c & = \argmax_{{\bf y} \in \mathcal{Y}} S^{k'}({\bf y}_c) \\
        \hat{\bf y}_a & = \argmax_{\mathbf{y} \in \mathcal{Y}} S^k(\mathbf{y}_c) \pm \epsilon \Delta(\mathbf{y}_p, \hat{\bf y}_c) \\
        \hat{\bf y}_b & = \argmax_{\mathbf{y} \in \mathcal{Y}} S^k(\mathbf{y}_c) \pm \epsilon \Delta(\hat{\bf y}_c, {\hat{\bf y}'}_c)
	\end{split}
\end{equation}
In our experiments, we fix the temperature parameter $\epsilon$ as, $\epsilon = +1$.

\paragraph{Intuition for the gradient computation:} We now present an intuitive explanation of the computation of gradient, as given in equation (\ref{eq13:gradCondNet}). For an input ${\bf x}$ and two noise samples $\mathbf{z}^k, \mathbf{z}^{k'}$, the conditional network outputs two scores $ S^{k}(\mathbf{y}_c)$ and $ S^{k'}(\mathbf{y}_c)$, with the corresponding maximum scoring outputs $\hat{\bf y}_c$ and ${\hat{\bf y}'}_c$. The model parameters $\boldsymbol{\theta}_c$ are updated via gradient descent in the negative direction of $\nabla_{\boldsymbol{\theta}_c} DISC_{\Delta}^{\epsilon}(\Pr\nolimits_p({\boldsymbol{\theta}}_p), \Pr\nolimits_c({\boldsymbol{\theta}}_c))$.
\begin{itemize}
    \item The term $DIV_{\Delta}^{\epsilon}(\Pr\nolimits_p, \Pr\nolimits_c)$ updates the model parameters towards the maximum scoring prediction $\hat{\bf y}_c$ of the score $ S^{k}(\mathbf{y}_c)$ while moving away from $\hat{\bf y}_a$, where $\hat{\bf y}_a$ is the sample corresponding to the maximum loss augmented score $S^{k}({\bf y}_c) \pm \epsilon \Delta(\mathbf{y}_p, \hat{\bf y}_c)$ with respect to the fixed prediction distribution samples ${\bf y}_p$. This encourages the model to move away from the prediction providing high loss with respect to the pseudo ground truth labels.
    \item The term $\gamma DIV_{\Delta}^{\epsilon}(\Pr\nolimits_c, \Pr\nolimits_c)$ updates the model towards $\hat{\bf y}_b$ and away from the $\hat{\bf y}_c$. Note the two negative signs giving the update in the positive direction. Here $\hat{\bf y}_b$ is the sample corresponding to the maximum loss augmented score $S^k(\mathbf{y}_c) \pm \epsilon \Delta(\hat{\bf y}_c, {\hat{\bf y}'}_c)$ with respect to the other prediction ${\hat{\bf y}'}_c$, encouraging diversity between $\hat{\bf y}_c$ and ${\hat{\bf y}'}_c$.
\end{itemize}


\begin{algorithm}[!t]
\RestyleAlgo{boxed}
\DontPrintSemicolon
\SetAlgoLined
\SetKwInOut{Input}{Input}\SetKwInOut{Output}{Output}

\Input{Training input $({x}, {a}) \in \mathcal{W}$, and prediction network outputs ${y}_p$}
\Output{$\hat{\bf y}_c^{1}, \dots, \hat{\bf y}_c^K$, sample $K$ predictions from the model}
\BlankLine
\For{$k = 1 \dots K$}
{
	Sample noise vector ${\bf z}^k$, generate output $\hat{\bf y}_c^k$:
    \[ \hat{\bf y}_c^k = \argmax_{{\bf y} \in \mathcal{Y}} S^k({\bf y}_c) \] 
    
    Find loss augmented prediction $\hat{\bf y}_a^k$ w.r.t. output from prediction network $\bf y_p$:
    \[ \hat{\bf y}_a^k = \argmax_{{\bf y} \in \mathcal{Y}} S^k({\bf y}_c) \pm \epsilon \Delta({\bf y}_p, \hat{\bf y}_c^k) \] 

}

Compute loss augmented predictions:\\
\For{$k = 1,\dots,K$}
{
	\For{$k' = 1,\dots,K, k' \neq k$}
    {
    	Find loss augmented prediction $\hat{\bf y}_b^k$ w.r.t. other conditional network outputs $\hat{\bf y}_c^k$:
        \[ \hat{\bf y}_b^{k,k'} = \argmax_{{\bf y} \in \mathcal{Y}} S^k({\bf y}_c) \pm \epsilon \Delta(\hat{\bf y}_c^k, {\hat{\bf y}'}_c) \] 
    }
}

Compute unbiased approximate gradients for $DIV_{\Delta}^{\epsilon}(\Pr\nolimits_c, \Pr\nolimits_c)$ and $DIV_{\Delta}^{\epsilon}(\Pr\nolimits_c, \Pr\nolimits_c)$ as:
\begin{dmath*}
	DIV_{\Delta}^{\epsilon}(\Pr\nolimits_p, \Pr\nolimits_c) = \frac{1}{K} \sum_{k=1}^K \Big[ \nabla_{\boldsymbol{\theta}_c} S^k(\hat{\bf y}_a) - \nabla_{\boldsymbol{\theta}_c} S^k(\hat{\bf y}_c) \Big], 
\end{dmath*} 
\begin{dmath*}
    DIV_{\Delta}^{\epsilon}(\Pr\nolimits_c, \Pr\nolimits_c) =  \frac{2}{K(K-1)} \sum_{\substack{k,k'=1\\k' \neq k}}^K \Big[ \nabla_{\boldsymbol{\theta}_c} S^k(\hat{\bf y}_b) - \nabla_{\boldsymbol{\theta}_c} S^{k'}({\hat{\bf y}'}_c) \Big]. 
\end{dmath*} 

Update model parameters by descending to the approximated gradients:
\[ \boldsymbol{\theta}_c^{t+1} = \boldsymbol{\theta}_c^{t} - \eta \nabla_{\boldsymbol{\theta}_c} DISC_{\Delta} (\Pr\nolimits_p(\boldsymbol{\theta}_p), \Pr\nolimits_c(\boldsymbol{\theta}_c)) \]

\caption{\emph{Conditional network training algorithm}}
\label{algo1:condNet}
\end{algorithm}

\paragraph{Training algorithm for conditional network:} Pseudo-code for training the conditional network for a single sample from weakly supervised data is presented in algorithm~\ref{algo1:condNet} below. In algorithm~\ref{algo1:condNet}, statements 1 to 3 describe the sampling process and computing the loss augmented prediction. We first sample $K$ different predictions $\hat{\bf y}_c^k$ corresponding to each noise vector ${\bf z}^k$ in statement 2. For the sampled prediction $\hat{\bf y}_c^k$ we compute the maximum loss augmented score $S^{k}(\mathbf{y}_c) \pm \epsilon \Delta(\mathbf{y}_p, \hat{\bf y}_c)$. This is then used to find the loss augmented prediction $\hat{\bf y}_a$ given in statement 3. 

In order to compute the gradients of the self diversity of conditional distribution, we need to find the maximum loss augmented prediction $\hat{\bf y}_b$. Here, the loss is computed between a pair of $K$ different predictions of the conditional network that we have already obtained. This is shown by statements 4 to 7 in algorithm~\ref{algo1:condNet}.

For the purpose of optimizing the conditional network using gradient descent, we need to find the gradients for the objective function of the conditional network defined in equation (\ref{cond_opt}) above. The computation of the unbiased approximate gradients for the individual terms in the objective function is shown in statement 8. We finally optimize the conditional network by the employing gradient descent step and updating the model parameters by descending to the approximated gradients as shown in statement 9 of algorithm~\ref{algo1:condNet}.

\section{Experiments}
In this section we present the implementation details of the prediction and the conditional network. Next, we present details of our ResNet based architecture and the detailed class-specific results on the Pascal VOC 2012 data set.

\subsection{Implementation Details}\label{ssec:implementation_details}
We use the standard Mask R-CNN as the prediction network and adapt the U-Net architecture for the conditional network, as shown figure~\ref{fig:cond_net}. For a fair comparison, the prediction network, we use ImageNet pre-trained ResNet-50 architecture for experiments with image-level annotation and a pretrained ResNet-101 architecture for the bounding box annotations.

Similar to~\cite{kohl2018probabilistic}, the U-Net architecture is modified by adding a noise sample as an extra channel after the deconvolutional layers as shown in figure~\ref{fig:cond_net}. A $1\times1$ convolution is applied to bring the number of channels back to the original dimensions ($32$ channels). The segmentation region proposal masks taken from MCG~\cite{APBMM2014} is then multiplied element-wise with the features from all the channels. This allows us to extract features only from the segmentation proposal. A $1\times1$ convolution is applied again to make the number of channels equal to the number of classes. This is followed by a global average pooling layer which gives us, for each of the segmentation proposals, a vector of dimensions equal to the number of classes. This vector for each of the segmentation proposal is passed to the inference algorithm which in turn provides the output segmentation masks corresponding to the image-level annotations. For all our experiments we choose K=10 for the conditional network and use the Adam optimizer. For all the other hyper-parameters we use the same configuration as described in~\cite{kohl2018probabilistic}. For the prediction network, we use default hyper-parameters described in~\cite{he2017mask}.

\subsection{ResNet based architecture for the conditional network}
In section 6.4 of the main paper, we study the effect of an alternative architecture for the conditional network. In what follows, we provide the details of this ResNet based conditional network.

\begin{figure}
    \centering
    \includegraphics[width=0.99\textwidth]{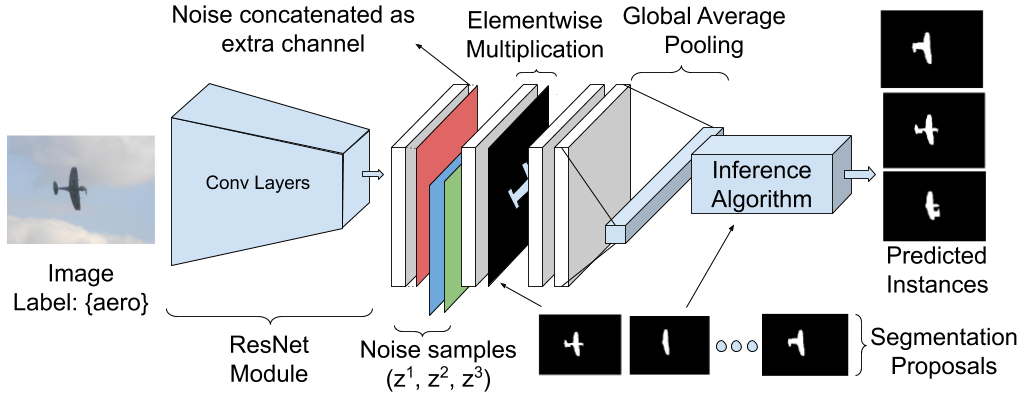}
    \caption{\emph{ResNet based conditional network}}
    \label{fig:res_condnet}
\end{figure}

The architecture for the ResNet based conditional network is shown in figure~\ref{fig:res_condnet}. The image is first passed through the ResNet module to obtain low-resolution high-level features. For the experiments where we use only the image-level annotations, a ResNet-50 module is employed and where we use the bounding-box level annotations, a ResNet-101 module is used. A noise filter is appended as an extra channel followed by a $1 \times 1$ convolutional filter, which brings the number of channels back to the original dimensions. The segmentation proposal masks are then multiplied element-wise to obtain segmentation proposal specific features. Next, a $1 \times 1$ convolutional is applied to make the number of channels equal to the number of classes. Finally, a global average pooling is applied to obtain a vector whose dimensions is equal to the number of classes in the data set. This vector is then passed through the inference algorithm to obtain the final predicted samples. As mentioned in section 6.4 of the paper, the results obtained using this ResNet based conditional network architecture are called as Ours (ResNet-50) and Ours (ResNet-101).

Note that, the U-Net based conditional network provides a higher resolution image features as compared to its ResNet based counterparts. These are then used to obtain the individual features of the segmentation mask proposals. The higher resolution features thus provide richer per-mask features. These are especially useful for smaller objects and cluttered environment where context resolution is important. The superior results of our method when using a U-Net based conditional network empirically verify this claim.

\subsection{Class specific results on VOC 2012 data set}
\begin{figure}[t]
    \centering
    \includegraphics[width=0.99\textwidth]{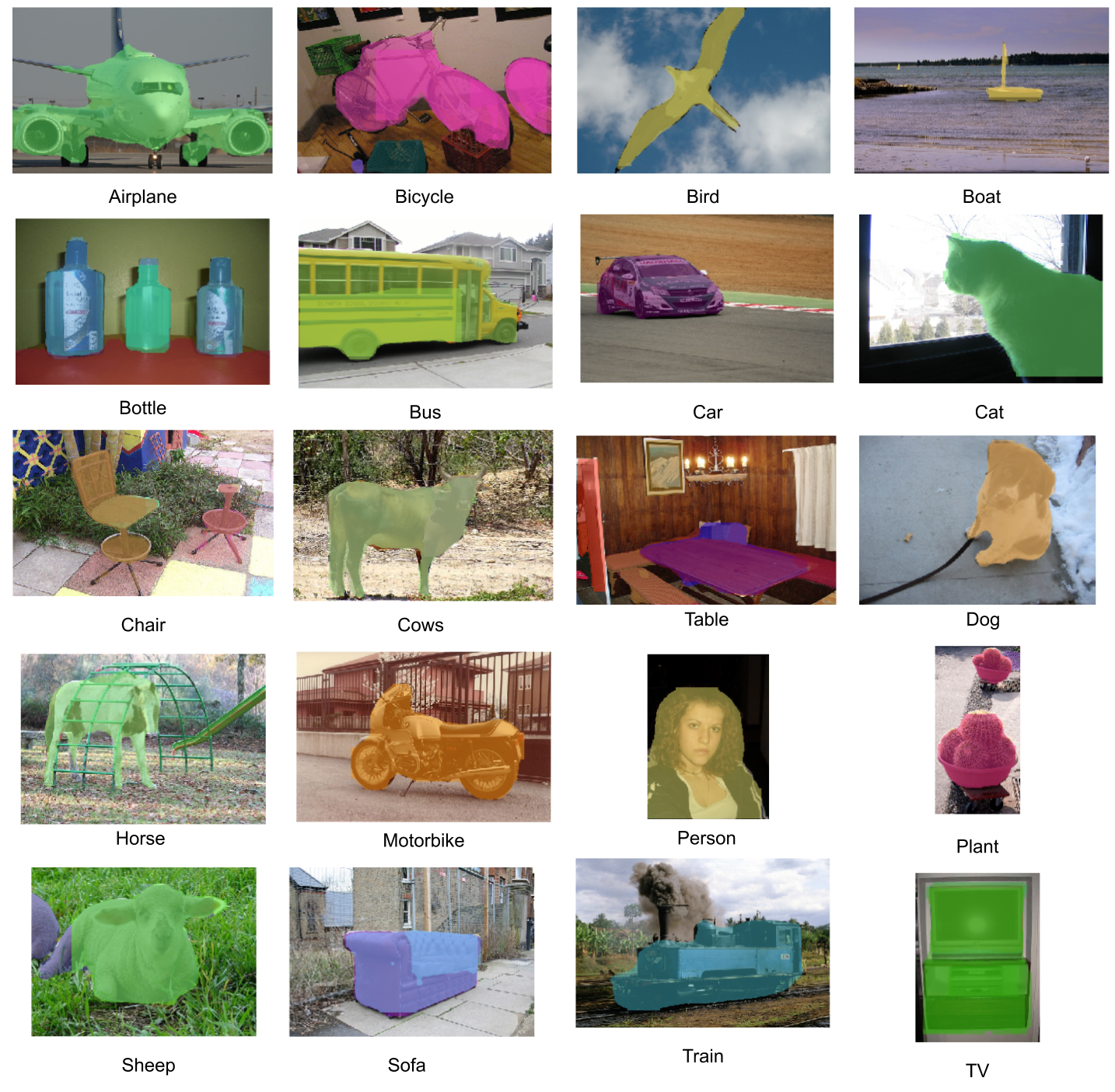}
    \caption{\emph{Qualitative results of our proposed approach on VOC 2012 validation set.}}
    \label{fig:qualitative results}
\end{figure}

We present the per-class result for our method on the Pascal VOC 2012 data set in table~\ref{tab:detailed_result}. The first two rows correspond to the result where our method was trained only using the image-level annotations. The last two rows correspond to the results where our methods were trained using the bounding box annotations.The qualitative results for each class is presented in figure~\ref{fig:qualitative results}.

\begin{table}[t]
\centering
\caption{Per class result for $\text{mAP}_{0.5}^r$ metric on Pascal VOC 2012 data set for methods that are trained on using image-level supervision $\mathcal{I}$ and bounding box annotations $\mathcal{B}$}
\resizebox{\textwidth}{!}{%
\begin{tabular}{|c|cccccccccccccccccccc|c|}
\hline
\multicolumn{1}{|c|}{Method} & \multicolumn{1}{c}{aero} & \multicolumn{1}{c}{bike} & \multicolumn{1}{c}{bird} & \multicolumn{1}{c}{boat} & \multicolumn{1}{c}{bottle} & \multicolumn{1}{c}{bus} & \multicolumn{1}{c}{car} & \multicolumn{1}{c}{cat} & \multicolumn{1}{c}{chair} & \multicolumn{1}{c}{cow} & \multicolumn{1}{c}{table} & \multicolumn{1}{c}{dog} & \multicolumn{1}{c}{horse} & \multicolumn{1}{c}{mbike} & \multicolumn{1}{c}{pson} & \multicolumn{1}{c}{plant} & \multicolumn{1}{c}{sheep} & \multicolumn{1}{c}{sofa} & \multicolumn{1}{c}{train} & \multicolumn{1}{c}{tv} & \multicolumn{1}{|c|}{mAP} \\ \hline
\begin{tabular}{@{}c@{}}Ours (ResNet-50) \\ $\mathcal{I}$ \end{tabular} & 74.2 & 52.6 & 68.6 & 44.1 & 25.0 & 63.4 & 35.9 & 72.6 & 18.2 & 47.1 & 24.6 & 63.5 & 53.7 & 67.3 & 40.9 & 29.4 & 42.8 & 39.6 & 69.5 & 61.2 & 49.7    \\ \hline
\begin{tabular}{@{}c@{}}Ours \\ $\mathcal{I}$ \end{tabular} & 75.5 & 53.6 & 69.9 & 45.3 & 26.7 & 64.3 & 37.4 & 73.7 & 19.3 & 48.7 & 25.3 & 64.6 & 55.0 & 68.3 & 42.1 & 30.8 & 44.2 & 40.5 & 70.6 & 62.2 & 50.9    \\ \hline \hline
\begin{tabular}{@{}c@{}}Ours (ResNet-101) \\ $\mathcal{B}$ \end{tabular} & 77.9 & 62.6 & 73.8 & 49.0 & 35.9 & 72.6 & 45.8 & 78.4 & 29.7 & 55.7 & 31.9 & 70.6 & 61.3 & 73.6 & 49.2 & 39.9 & 50.8 & 47.9 & 76.5 & 69.6 & 57.7    \\ \hline
\begin{tabular}{@{}c@{}}Ours \\ $\mathcal{B}$ \end{tabular} & 79.1 & 63.9 & 75.1 & 49.3 & 36.5 & 73.1 & 46.4 & 78.8 & 30.1 & 56.4 & 32.1 & 71.3 & 61.6 & 74.8 & 49.5 & 40.2 & 51.1 & 48.3 & 77.2 & 69.9 & 58.2    \\ \hline
\end{tabular}
}
\label{tab:detailed_result}
\end{table}


\end{document}